\documentclass[]{arxiv_preprint}

\usepackage{amsmath}
\usepackage{amsfonts}
\usepackage{amssymb}
\usepackage{xspace}
\usepackage{wrapfig}

\newcommand{\ourmethod}{DyPES-VLA\xspace}

\newcommand{\subrow}{\hspace{1.1em}}

\title{\textcolor{metablue}{DyPES-VLA}: Learning Shared Dynamics Priors and Embodiment-Specific Control for Cross-Embodiment Manipulation}

\author[1,\ast]{Junfeng Li}
\author[1,\ast]{Junjie He}
\author[1,\ast,\dagger]{Zhide Zhong}
\author[1,\ast]{Yangyang Zheng}
\author[2]{Pingyue Sheng}
\author[2]{Jiayu Dong}
\author[1]{Ruixin Li}
\author[1]{Haodong Yan}
\author[1]{Jiaguan Zhu}
\author[1]{Tianran Zhang}
\author[1]{Runze Yu}
\author[1]{Wen Chen}
\author[1]{Liuqing Yang}
\author[2]{Yuxiang Gao}
\author[1,\ddagger]{Haoang Li}

\affiliation[1]{The Hong Kong University of Science and Technology (Guangzhou), Guangzhou, China}
\affiliation[2]{COCO Matrix, Shanghai, China}
\contribution[\ast]{Equal contribution}
\contribution[\dagger]{Project Leader}
\contribution[\ddagger]{Corresponding author.}

\abstract{
  Vision-Language-Action (VLA) models have become a powerful paradigm for robot manipulation, but training a single generalist policy for heterogeneous robot embodiments remains an open problem.
Existing methods have two main limitations. First, they underuse dynamics priors shared across diverse visual and interaction data, limiting cross-embodiment transfer. Second, they require extensive manual preprocessing to convert embodiment-specific actions into a common format.
To overcome these limitations, we propose \ourmethod, a cross-embodiment VLA that learns shared \textbf{Dy}namics \textbf{P}riors and \textbf{E}mbodiment-\textbf{S}pecific control.
First, we learn shared dynamics priors by training the vision-language model (VLM) with a future-prediction objective on cross-embodiment data, driving the shared query representation to capture object motion, contact, and interaction-induced scene changes.
Second, an embodiment-specific Mixture-of-Experts (MoE) action head translates these shared dynamics priors into executable controls directly in each embodiment's native action space, without manually pre-aligning heterogeneous actions into a common format.
This head shares attention layers to capture common temporal action structures, while its embodiment-specific feed-forward experts resolve the unique kinematic constraints and control semantics of distinct embodiments.
As a generalist policy, our \ourmethod achieves state-of-the-art performance across simulation and real-world evaluations, reaching 98.0\% success on LIBERO, 59.25\% on RoboCasa-GR1, and 89.02\% on RoboTwin~2.0.

}

\metadata[Project Page]{\url{https://livfour.github.io/DyPES-VLA_RELEASE/}}

\begin{document}

\maketitle

\section{Introduction}
\label{sec:intro}

Vision-Language-Action (VLA) models have achieved strong performance across a range of robotic manipulation tasks~\cite{brohan2023rt,zitkovich2023rt,kim2024openvla,octo,BlackK-RSS-25,bjorck2025gr00t,rdt1b}.
However, most VLA policies remain tied to a specific robot and struggle to generalize across embodiments~\cite{zhangdreamvla}.
Recent efforts~\cite{openxembodiment,univla,zheng2025x,xdiffvla,himoevla} train VLAs on heterogeneous cross-embodiment data to move toward generalist robot manipulation.
The central challenge of training is to determine what knowledge should be shared across embodiments and what must remain embodiment-specific.

Despite this progress, two limitations remain in existing cross-embodiment VLAs.
First, existing methods rely on action prediction as their only supervision signal, so shared dynamics priors are learned only from action labels. This underuses large collections of human and robot manipulation videos, despite their recurring patterns of object motion, contact, and scene evolution.
Second, many existing approaches share information at the action level by manually mapping heterogeneous controls into a common format~\cite{egovla,beingh0,rdt1b}. Although this preprocessing captures coarse trajectories, it requires coordinate transformations or inverse kinematics and scales poorly across robot morphologies. More fundamentally, forcing heterogeneous robots into a common action format entangles two things that should stay separate: the interaction regularities shared across embodiments and the control semantics unique to each robot's body.

\begin{wrapfigure}{r}{0.48\textwidth}
    \centering
    \includegraphics[width=\linewidth]{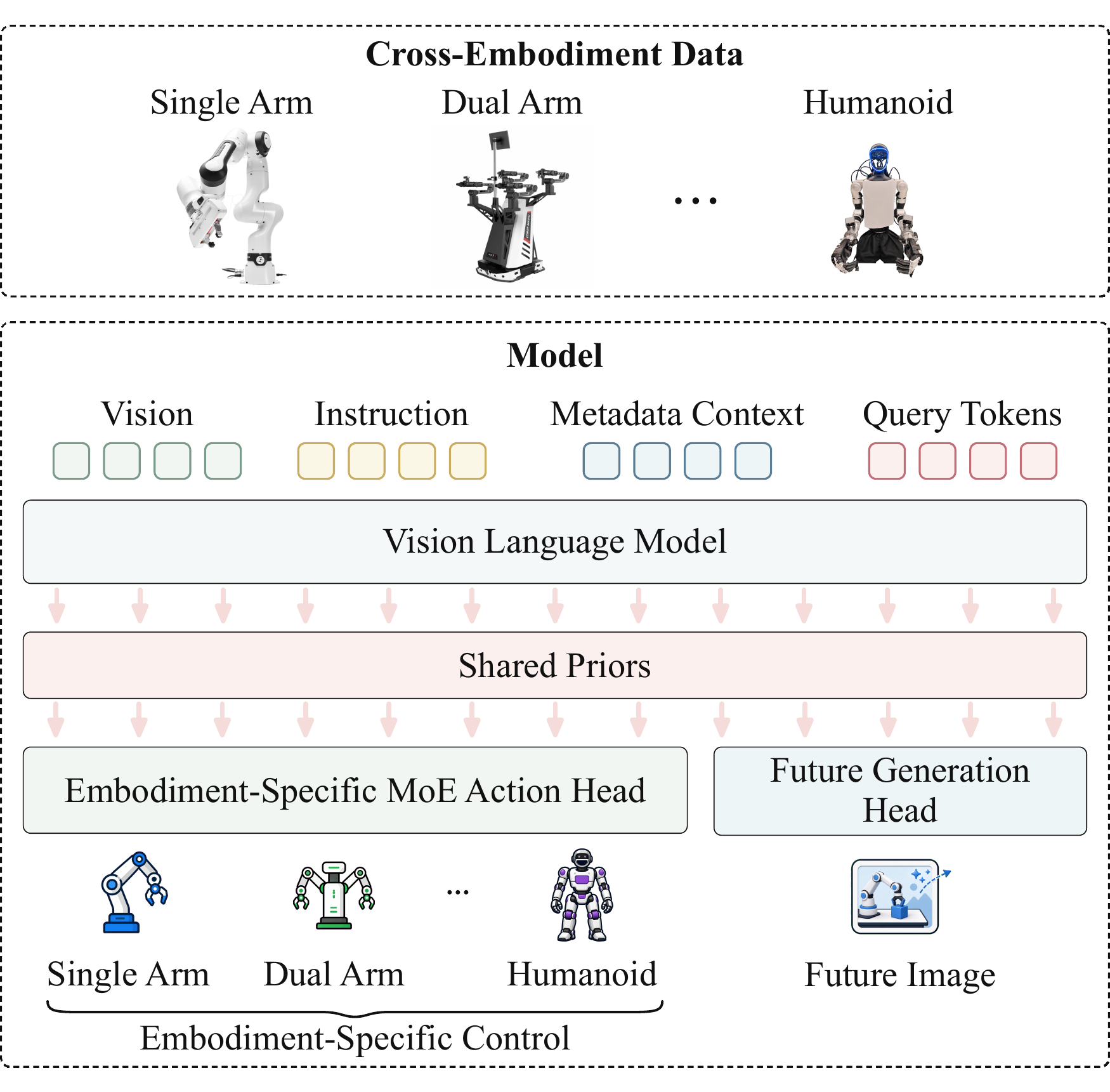}
    \caption{
        We propose a paradigm that learns from heterogeneous cross-embodiment data.
        Our \ourmethod learns shared dynamics priors and embodiment-specific control.
        The former is supervised by a future generation head, and the latter is achieved by an MoE action head.
    }
    \label{fig:teaser}
\end{wrapfigure}

To address these limitations, we propose a cross-embodiment learning paradigm for generalist manipulation that learns shared dynamics priors and embodiment-specific control, and instantiate it as our \ourmethod.
In principle, this paradigm admits data from arbitrary embodiments for joint training without additional action-space alignment.
In this paper, we instantiate it on three embodiment families, each spanning a simulated and a physical robot: a single-arm platform (Franka Emika Panda and Franka Research 3), a dual-arm platform (ALOHA-AgileX and AgileX Robotics COBOT Magic), and a humanoid platform (Fourier GR-1 and Unitree G1 with Inspire RH56DFQ hands).
As illustrated in Fig.~\ref{fig:teaser}, our \ourmethod learns shared dynamics priors through future-prediction supervision and translates the resulting predictive query representation into embodiment-specific actions through the MoE action head. Unlike world-action models (WAMs) that couple future prediction with action generation~\cite{uwm,yuan2026fastwam}, our \ourmethod uses future prediction only to learn shared dynamics priors, leaving action generation to a dedicated embodiment-specific head.

\begin{figure*}[t!]
\centering
\includegraphics[width=1.0\linewidth]{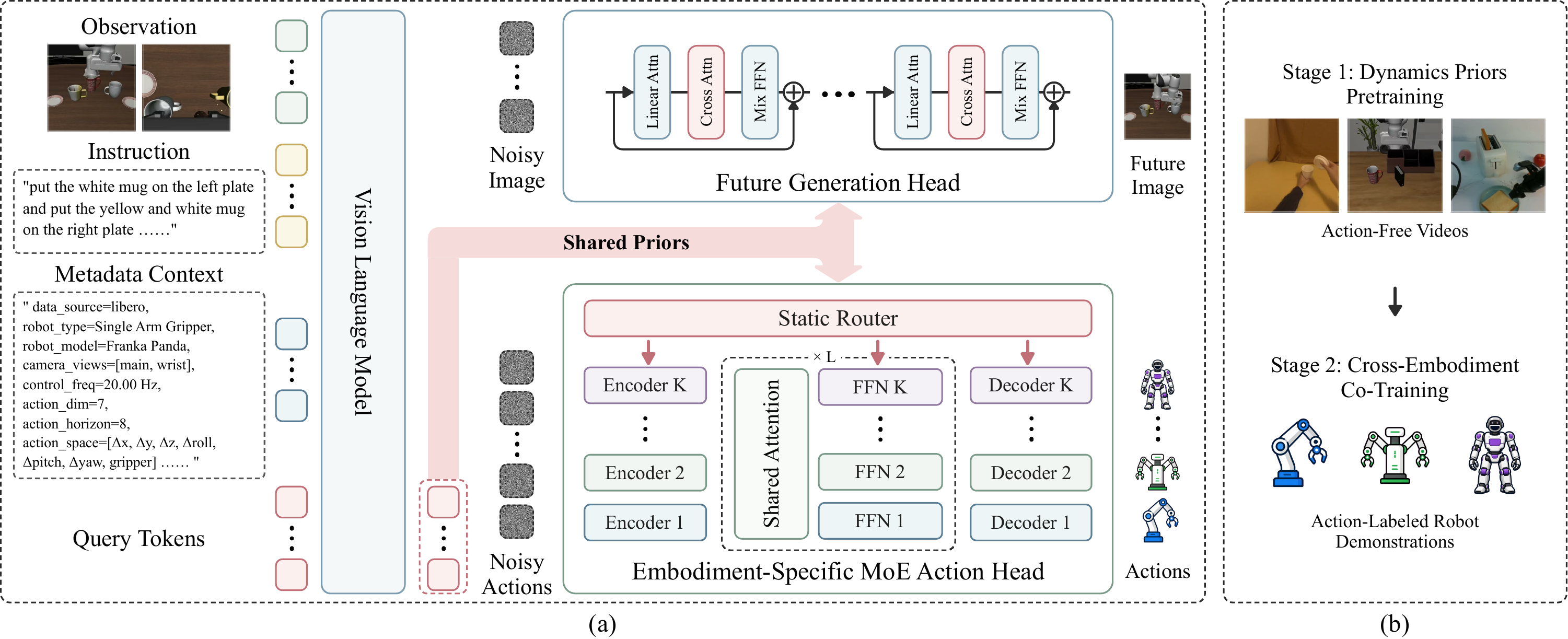}
    \caption{Overview of our \ourmethod.
    (a) A pretrained Vision Language Model maps observations, instructions, embodiment metadata, and learnable query tokens into query states that carry the shared dynamics priors.
    A future generation head predicts the future frame from these states, and an embodiment-specific Mixture-of-Experts (MoE) action head decodes them into actions in each embodiment's native action space.
    The action head shares attention layers across embodiments, while a static router selects the embodiment-specific encoder, feed-forward expert, and decoder.
    (b) Two-stage training: Dynamics Priors Pretraining on action-free videos, followed by Cross-Embodiment Co-Training on action-labeled robot demonstrations.}
    \label{fig:architecture}
\end{figure*}

Specifically, as shown in Fig.~\ref{fig:architecture}, we use a pretrained vision-language model (VLM) to map visual observations, language instructions, embodiment metadata, and learnable query tokens into query states that form a shared interface between future prediction and embodiment-specific control. We train our \ourmethod in two stages. In the first stage, we optimize the VLM, query tokens, and future generation head on action-free human and robot videos by predicting future frames. Through this future-prediction supervision, we encourage the query states to retain information about object motion, contact, and interaction-induced changes; we refer to these predictive regularities as shared dynamics priors. In the second stage, we jointly optimize the future-prediction and action objectives on action-labeled demonstrations from multiple embodiments. We continue to use future prediction to regularize the shared query representation, while we condition an embodiment-specific MoE action head on the same representation to directly generate action chunks in the native action space of each robot. In this way, we first learn dynamics priors shared across embodiments and then translate them into embodiment-specific control.
We evaluate our \ourmethod with a single co-trained checkpoint on three simulation benchmarks. It achieves 98.0\% success on LIBERO~\cite{liu2023libero}, 59.25\% on RoboCasa-GR1~\cite{robocasa,bjorck2025gr00t}, and 89.02\% on RoboTwin~2.0~\cite{robotwin,robotwin2}. We then jointly finetune the same checkpoint on demonstrations from three physical embodiments. The resulting unified policy averages 75.6\% success on three tasks across three real-world embodiments.

\noindent In summary, our main contributions are:
\begin{itemize}
    \item We propose a cross-embodiment learning paradigm for generalist robot manipulation that learns shared dynamics priors and embodiment-specific control.
    
    \item We introduce our \ourmethod, an effective instantiation of this paradigm that learns shared priors through future-supervised query states and realizes control through an embodiment-specific action head, without requiring a common action format.
    
    \item Extensive experiments demonstrate that our \ourmethod achieves state-of-the-art performance across three simulation benchmarks in a single-checkpoint setting. Real-world experiments show that the dynamics priors learned through cross-embodiment co-training provide a transferable foundation for physical control across distinct morphologies and action spaces.
\end{itemize}

\section{Related Work}
\label{sec:related}

\paragraph{Vision-Language-Action Models.}
Built on pretrained vision-language models, Vision-Language-Action (VLA) models and related video-action models achieve strong performance in robotic manipulation~\cite{brohan2023rt,zitkovich2023rt,kim2024openvla,song2026reconvla,li2026spatial,song2025pd,chen2026unified,yan2026svam}.
Early VLAs are typically specialized to a particular robot, and deployment on a new embodiment commonly requires embodiment-specific finetuning~\cite{KimM1-RSS-25}.
Cross-robot corpora~\cite{openxembodiment} have since supported increasingly general policies such as Octo~\cite{octo} and the $\pi$ series~\cite{BlackK-RSS-25,PertschK-RSS-25,black2025pi}.
These efforts demonstrate the value of heterogeneous robot data, but leave open a question of what a single policy should share across embodiments, and what must remain embodiment-specific.

\paragraph{Cross-Embodiment VLA Models.}
Methods explicitly targeting heterogeneous embodiments address this question differently, mainly in terms of where embodiment specificity enters the model.
A first line unifies heterogeneous embodiments into a common action space before learning. Hand-engineered variants build this space with a shared end-effector frame~\cite{rdt2} or interpretable action vectors~\cite{rdt1b}, while learned variants map embodiments into a latent or universal action space~\cite{lapa,moto,univla,uniact}.
But such a common action space mixes what is shared across robots with what is specific to each robot.
A second line keeps a largely shared policy and supplies embodiment context separately. This context appears as soft prompts~\cite{zheng2025x}, motion-transfer training~\cite{gemini15}, or a single action head shared across embodiments~\cite{xdiffvla}.
A third line partitions the architecture itself, adding embodiment-specific modules to a shared backbone. These include per-embodiment stems and heads~\cite{hpt}, readouts~\cite{octo,crossformer}, state encoders and action decoders~\cite{bjorck2025gr00t}, and experts routed over heterogeneity factors~\cite{himoevla}.
Our \ourmethod follows this architectural-specialization direction but moves the shared interface upstream to future-supervised query states, which embodiment-specific experts decode into native action spaces.

\paragraph{Predictive Learning for Cross-Embodiment Manipulation.}
Future prediction provides a unified supervision signal that applies to cross-embodiment data, and existing work exploits it along two lines.
One line couples prediction with control. World-action models (WAMs) jointly model future visual and action streams~\cite{pad,uva,worldvla,dit4dit,motus}, and some VLA models further use predicted futures to guide action generation and planning~\cite{zhangdreamvla}.
Another line uses prediction to shape the policy representation. Such methods learn from visual future prediction~\cite{gr1,vpp,genieenvisioner,zhong2025flowvla}, latent future alignment~\cite{flare,zhong2026dualcotvla}, and jointly trained future and action branches~\cite{seer,zhao2025cot,lv2025f1,mantis,wla0}.
Closest to our work, UWM couples video and action diffusion to incorporate action-free videos~\cite{uwm}. LDA-1B scales joint dynamics, forecasting, and policy learning over heterogeneous data~\cite{lda1b}. Fast-WAM removes explicit future synthesis at inference to isolate the benefit of video co-training~\cite{yuan2026fastwam}.
Our \ourmethod instead uses future-supervised query states to learn shared dynamics priors and decodes them into native controls through embodiment-specific experts.

\section{Methodology}
\label{sec:method}
\subsection{Method Overview}
\label{subsec:method_overview}

As shown in Fig.~\ref{fig:architecture}, our \ourmethod comprises three components that jointly learn shared dynamics priors and embodiment-specific control.
\textbf{1)} A pretrained VLM~\cite{bai2025qwen3} encodes the visual observations, the instruction, and the embodiment metadata, together with a set of learnable query tokens appended to the multimodal sequence; the output query states form the shared interface between the two heads.
\textbf{2)} A future generation head, instantiated as a SANA image generator~\cite{xie2024sana}, is conditioned on these query states and trained to synthesize the future frame, driving the shared query representation to capture the dynamics priors of object motion, contact, and interaction-induced scene changes.
\textbf{3)} An embodiment-specific MoE action head, instantiated as a flow-matching Diffusion Transformer (DiT)~\cite{peebles2023scalable}, is conditioned on the same query states. It then generates action chunks in each embodiment's native action space.
Training proceeds in two stages: the VLM, the query tokens, and the future generation head are first pretrained with the future-prediction objective on large-scale action-free videos, and both heads are then jointly optimized on robot demonstrations; the future generation head is removed at inference.

\subsection{Shared Query Interface}
\label{subsec:method_query}

We bridge multimodal understanding and downstream generation with a single set of learnable query tokens $\mathbf{Q}\in\mathbb{R}^{N\times d_{\textnormal{VLM}}}$.
The embodiment metadata $m_{e}$ is verbalized into a compact textual context (e.g., \texttt{data\_source}, \texttt{robot\_type}, and \texttt{control\_freq}), making the current embodiment explicit to the VLM.
The query tokens are appended to the encoded multimodal sequence and processed by the VLM $f_{\theta}$ in one forward pass:
\begin{equation}
    \mathbf{Z}
    =
    f_{\theta}\big(
    \left[
    \phi_{v}(\mathbf{o}_{t}),\,
    \phi_{\ell}(\ell),\,
    \phi_{\ell}(m_{e}),\,
    \mathbf{Q}
    \right]
    \big),
    \label{eq:backbone}
\end{equation}
where $\phi_{v}$ and $\phi_{\ell}$ denote the visual and language tokenizers, and $\mathbf{Z}\in\mathbb{R}^{N\times d_{\textnormal{VLM}}}$ collects the last-layer hidden states of the query tokens.
We refer to $\mathbf{Z}$ as the query states.
The policy takes no proprioceptive input.

\subsection{Learning Dynamics Priors}
\label{subsec:method_future}

The future generation head supervises $\mathbf{Z}$ with a generative future-prediction objective shared by all embodiments and data sources.
Given the current context at time $t$, the target is the future frame $\mathbf{x}_{t+\Delta_{e}}$ from the primary camera, where the temporal offset $\Delta_{e}$ is set per embodiment to match its action horizon $H_{e}$.

A frozen autoencoder $\textnormal{AE}_{w}$~\cite{xie2024sana} encodes the future frame into a compact latent representation $\mathbf{z}=\textnormal{AE}_{w}(\mathbf{x}_{t+\Delta_{e}})$.
The query states are projected into the conditioning space of the SANA transformer $g_{\psi}$ by a lightweight projector $p_{\omega}$, and serve as its conditioning tokens via cross-attention.
The generator is conditioned only on the projected query states: it receives neither the current observation nor its latent representation, so information needed to synthesize the future must pass through the $N$ query states.

The head is trained with a rectified-flow objective~\cite{lipman2022flow}: for a noise sample $\boldsymbol{\epsilon}\sim\mathcal{N}(\mathbf{0},\mathbf{I})$ and noise level $\tau\sim\mathcal{U}(0,1)$, the noisy latent representation is $\mathbf{z}_{\tau}=\tau\mathbf{z}+(1-\tau)\boldsymbol{\epsilon}$ and the head regresses the straight-line velocity toward the data:
\begin{equation}
    \mathcal{L}_{\textnormal{future}}
    =
    \mathbb{E}_{\tau,\boldsymbol{\epsilon}}
    \Big[
    \big\|
    g_{\psi}\big(\mathbf{z}_{\tau},\tau,p_{\omega}(\mathbf{Z})\big)
    -
    (\mathbf{z}-\boldsymbol{\epsilon})
    \big\|_{2}^{2}
    \Big].
    \label{eq:future_loss}
\end{equation}

\subsection{Learning Embodiment-Specific Control}
\label{subsec:method_moe}

The action head translates the shared predictive query representation $\mathbf{Z}$ into executable controls.
It is a flow-matching DiT that factorizes action generation into two parts: attention layers provide temporal computation shared across embodiments, while statically routed experts handle embodiment-specific realization.
Embodiment metadata $m_{e}$ deterministically assigns each sample a routing index $r(e)\in\{1,\dots,K\}$, where $K$ is the number of supported embodiments.

\paragraph{Per-Embodiment Interfaces.}
Each embodiment owns a lightweight encoder--decoder pair that adapts its native action space to the shared DiT width.
The encoder $\textnormal{Enc}_{r(e)}$ embeds the noisy action chunk together with the flow timestep.
The decoder $\textnormal{Dec}_{r(e)}$ maps DiT outputs back to a velocity over the native action chunk $\mathbf{A}^{e}\in\mathbb{R}^{H_{e}\times d_{e}}$.

\paragraph{Shared Attention, Routed Experts.}
The DiT stacks $L$ transformer blocks that apply cross-attention to the query states $\mathbf{Z}$ and self-attention over the action sequence, both modulated by adaptive layer normalization (AdaLN) conditioned on the flow timestep $\tau$.
Within every block, the attention operation (Attn) is shared across all embodiments, while the feed-forward network (FFN) is a bank of $K$ experts selected by the routing index:
\begin{equation}
\begin{aligned}
    \bar{\mathbf{X}}
    &=
    \mathbf{X}
    +
    \textnormal{Attn}\big(\textnormal{AdaLN}(\mathbf{X},\tau);\,\mathbf{Z}\big),\\
    \mathbf{X}'
    &=
    \bar{\mathbf{X}}
    +
    \textnormal{FFN}^{(r(e))}\big(\textnormal{AdaLN}(\bar{\mathbf{X}},\tau)\big),
\end{aligned}
\label{eq:moe_block}
\end{equation}
where $\mathbf{X}$ denotes the action token sequence.

The head is trained with the same rectified-flow formulation as the future generation head.
The timestep is drawn from the Beta schedule $\tau{=}s(1{-}u)$ with $u\sim\textnormal{Beta}(1.5,1.0)$ and $s{=}0.999$, which places most of the sampling mass near the high-noise end $\tau{=}0$~\cite{bjorck2025gr00t}.
With noise $\boldsymbol{\epsilon}\sim\mathcal{N}(\mathbf{0},\mathbf{I})$, the noisy chunk is $\mathbf{A}^{e}_{\tau}=\tau\mathbf{A}^{e}+(1-\tau)\boldsymbol{\epsilon}$, and the head predicts the velocity
\begin{equation}
    \hat{\mathbf{V}}
    =
    \textnormal{Dec}_{r(e)}\Big(
    \textnormal{MoEDiT}\big(
    \textnormal{Enc}_{r(e)}(\mathbf{A}^{e}_{\tau},\tau),\,
    \mathbf{Z},\,
    \tau,\,
    r(e)
    \big)\Big),
    \label{eq:action_pred}
\end{equation}
supervised by a regression loss in each embodiment's native action space:
\begin{equation}
    \mathcal{L}_{\textnormal{action}}
    =
    \mathbb{E}_{\tau,\boldsymbol{\epsilon}}
    \Big[
    \big\|
    \hat{\mathbf{V}}
    -
    (\mathbf{A}^{e}-\boldsymbol{\epsilon})
    \big\|_{2}^{2}
    \Big].
    \label{eq:action_loss}
\end{equation}

\subsection{Two-Stage Training and Inference}
\label{subsec:method_training}

\paragraph{Stage 1: Dynamics Priors Pretraining on Action-Free Videos.}
Since the future-prediction objective in Eq.~\eqref{eq:future_loss} requires no action labels, the first stage pretrains the VLM, the query tokens, and the SANA head on large-scale action-free videos.
We draw egocentric human manipulation videos from EgoDex~\cite{hoque2025egodex}, together with simulation videos from the embodiments used in co-training.

\paragraph{Stage 2: Cross-Embodiment Co-Training on Action-Labeled Demonstrations.}
The second stage trains on action-labeled robot demonstrations from multiple embodiments and optimizes both heads jointly:
\begin{equation}
    \mathcal{L}
    =
    \mathcal{L}_{\textnormal{action}}
    +
    \lambda_{w}\,
    \mathcal{L}_{\textnormal{future}},
    \label{eq:total_loss}
\end{equation}
where $\lambda_{w}$ balances future prediction against action learning.

\paragraph{Inference.}
At deployment, one VLM forward pass produces $\mathbf{Z}$. 
Conditioned on $\mathbf{Z}$, the action head integrates the learned flow from Gaussian noise with a few Euler steps, outputting a native action chunk of the corresponding embodiment.
The future generation head is skipped at inference.

\section{Experiments}
\label{sec:exp}

In this section, we evaluate whether future-supervised dynamics priors and the embodiment-specific MoE action head provide an effective paradigm for cross-embodiment manipulation learning.
We organize the experiments around the following questions:

\begin{itemize}
    \item \textbf{Q1: Can our \ourmethod use a single checkpoint to perform across simulation benchmarks with distinct embodiments and native action spaces?}
    \item \textbf{Q2: Does future supervision enrich the shared representation with dynamics priors and improve cross-embodiment policy learning?}
    \item \textbf{Q3: Does embodiment-specific action realization mitigate interference from heterogeneous control spaces?}
    \item \textbf{Q4: Does embodiment metadata resolve embodiment and data-source ambiguity?}
\end{itemize}

\subsection{Experimental Setup}
\label{sec:exp_setup}

\paragraph{Benchmarks.}
We evaluate our \ourmethod on three simulation benchmarks and three real-world robot platforms, each instantiated with a specific robot embodiment.
\begin{itemize}
    \item \textbf{RoboTwin~2.0}~\cite{robotwin,robotwin2} evaluates multi-task manipulation over its 50 tasks with a 14-DoF ALOHA-AgileX dual-arm robot.
    \item \textbf{RoboCasa-GR1}~\cite{robocasa,bjorck2025gr00t} evaluates kitchen manipulation with a 29-DoF Fourier GR-1 humanoid.
    \item \textbf{LIBERO}~\cite{liu2023libero} evaluates tabletop manipulation with a simulated 7-DoF Franka Emika Panda single-arm robot.
    \item \textbf{Real-world platforms.}
    We further evaluate our \ourmethod on three physical embodiments that span distinct morphologies: a 7-DoF Franka Research 3 (FR3) single-arm robot, a 14-DoF AgileX Robotics COBOT Magic dual-arm robot, and a Unitree G1 humanoid equipped with Inspire RH56DFQ hands.
\end{itemize}

\begin{table}[t!]
\centering
\caption{
Results on the \textbf{RoboTwin~2.0} simulation benchmark~\cite{robotwin,robotwin2} with a 14-DoF dual-arm robot. Clean and Randomized correspond to the benchmark's Easy and Hard evaluation settings, respectively. Setting-wise results are reported by the cited papers, and Average is the arithmetic mean over both settings (\%).
\textbf{Bold} indicates the best result, and \underline{underline} indicates the second best.
}
\label{tab:main_robotwin}
\begingroup
\small
\setlength{\tabcolsep}{2.5pt}
\begin{tabular*}{\textwidth}{@{\extracolsep{\fill}}lccc@{}}
\toprule
Method & Clean & Randomized & Average \\
\midrule
\multicolumn{4}{@{}l}{\textit{Per-benchmark specialists}} \\
\subrow Diffusion Policy~\cite{chi2025diffusion} & 28.0 & 0.6 & 14.30 \\
\subrow RDT-1B~\cite{rdt1b} & 34.5 & 13.7 & 24.10 \\
\subrow $\pi_0$~\cite{BlackK-RSS-25} & 46.4 & 16.3 & 31.35 \\
\subrow X-VLA~\cite{zheng2025x} & 70.0 & 39.0 & 54.50 \\
\subrow $\pi_{0.5}$~\cite{black2025pi} & 82.7 & 76.8 & 79.75 \\
\subrow ABot-M0~\cite{abotm0} & 86.0 & 85.0 & 85.50 \\
\midrule
\multicolumn{4}{@{}l}{\textit{Single-checkpoint generalists}} \\
\subrow Qwen-VLA~\cite{qwenvla2026} & \underline{86.1} & \underline{87.2} & \underline{86.65} \\
\subrow \textbf{\ourmethod (ours)} & \textbf{88.78} & \textbf{89.26} & \textbf{89.02} \\
\bottomrule
\end{tabular*}
\endgroup

\vspace{0.9\baselineskip}

\centering
\caption{
Results on the \textbf{RoboCasa-GR1} simulation benchmark~\cite{robocasa,bjorck2025gr00t} with a 29-DoF humanoid, averaged over 50 rollouts per task (\%).
\textbf{Bold} indicates the best result, and \underline{underline} indicates the second best.
}
\label{tab:main_robocasa}
\begingroup
\small
\setlength{\tabcolsep}{4pt}
\begin{tabular*}{\textwidth}{@{\extracolsep{\fill}}lc@{}}
\toprule
Method & Success Rate (\%) \\
\midrule
\multicolumn{2}{@{}l}{\textit{Per-benchmark specialists}} \\
\subrow Diffusion Policy~\cite{chi2025diffusion} & 40.9 \\
\subrow GR00T-N1.5~\cite{bjorck2025gr00t} & 48.2 \\
\subrow GR00T-N1.6~\cite{bjorck2025gr00t} & 47.6 \\
\subrow Qwen3GR00T~\cite{starvla2025} & 47.8 \\
\subrow Qwen3PI~\cite{starvla2025} & 43.9 \\
\subrow Qwen3OFT~\cite{starvla2025} & 48.8 \\
\subrow Qwen3FAST~\cite{starvla2025} & 39.0 \\
\subrow LDA-1B~\cite{lda1b} & 55.4 \\
\subrow ABot-M0~\cite{abotm0} & \underline{58.3} \\
\midrule
\multicolumn{2}{@{}l}{\textit{Single-checkpoint generalists}} \\
\subrow Qwen-VLA~\cite{qwenvla2026} & 56.7 \\
\subrow \textbf{\ourmethod (ours)} & \textbf{59.25} \\
\bottomrule
\end{tabular*}
\endgroup
\end{table}

\begin{table}[t!]
\centering
\caption{
Per-suite success rates on \textbf{LIBERO}~\cite{liu2023libero} (\%). Results are reported by the cited papers; ``--'' denotes a breakdown not reported in the source paper. \textbf{Bold} indicates the best result, and \underline{underline} indicates the second best.
}
\label{tab:main_libero}
\begingroup
\scriptsize
\setlength{\tabcolsep}{1.5pt}
\begin{tabular*}{\textwidth}{@{\extracolsep{\fill}}lccccc@{}}
\toprule
Method & Spatial & Object & Goal & Long & Average \\
\midrule
\multicolumn{6}{@{}l}{\textit{Per-benchmark specialists}} \\
\subrow Diffusion Policy~\cite{chi2025diffusion} & 78.3 & 92.5 & 68.3 & 50.5 & 72.4 \\
\subrow OpenVLA~\cite{kim2024openvla} & 84.7 & 88.4 & 79.2 & 53.7 & 76.5 \\
\subrow $\pi_0$~\cite{BlackK-RSS-25} & 96.8 & 98.8 & 95.8 & 85.2 & 94.2 \\
\subrow $\pi_{0.5}$~\cite{black2025pi} & \textbf{98.8} & 98.2 & \textbf{98.0} & 92.4 & 96.9 \\
\subrow OpenVLA-OFT~\cite{KimM1-RSS-25} & 97.6 & 98.4 & \underline{97.9} & 94.5 & 97.1 \\
\subrow Fast-WAM~\cite{yuan2026fastwam} & \underline{98.2} & \textbf{100.0} & 97.0 & 95.2 & 97.6 \\
\subrow X-VLA~\cite{zheng2025x} & \underline{98.2} & 98.6 & 97.8 & \textbf{97.6} & \textbf{98.1} \\
\midrule
\multicolumn{6}{@{}l}{\textit{Single-checkpoint generalists}} \\
\subrow Qwen-VLA~\cite{qwenvla2026} & -- & -- & -- & -- & 97.9 \\
\subrow \textbf{\ourmethod (ours)} & \textbf{98.8} & \underline{99.4} & 97.0 & \underline{96.8} & \underline{98.0} \\
\bottomrule
\end{tabular*}
\endgroup
\end{table}

\paragraph{Training Data.}
Stage~1 trains the VLM, the query tokens, and the future generation head with the future-prediction objective on action-free videos from EgoDex~\cite{hoque2025egodex} and from the three simulation benchmarks, namely RoboTwin~2.0, RoboCasa-GR1, and LIBERO.
We use the full EgoDex corpus and sample this video mixture as 50\% EgoDex, 20\% RoboTwin~2.0, 20\% RoboCasa-GR1, and 10\% LIBERO.
Stage~2 co-trains both heads on action-labeled demonstrations from the same three benchmarks, mixed as 40\% RoboTwin~2.0, 40\% RoboCasa-GR1, and 20\% LIBERO.

\begin{figure*}[!t]
\centering
\includegraphics[width=\textwidth]{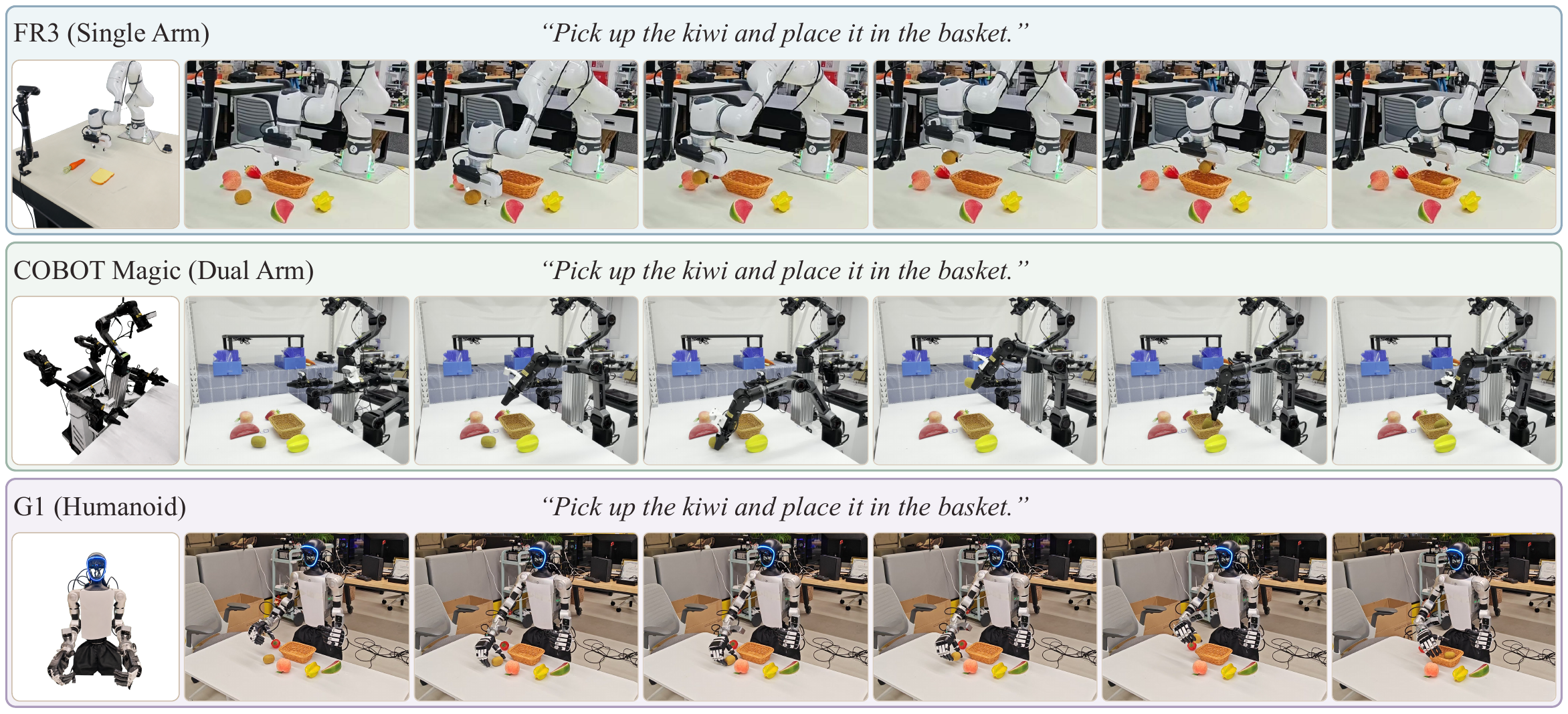}

\caption{
Real-world rollouts of our \ourmethod using a single checkpoint jointly finetuned on demonstrations from three physical embodiments with distinct morphologies.
Each row shows the hardware setup followed by six keyframes (left to right) of one representative task per platform: an FR3 single-arm robot (top), a COBOT Magic dual-arm robot (middle), and a G1 humanoid (bottom).
}
\label{fig:real_world}

\vspace{0.4\baselineskip}
\begin{minipage}{\textwidth}
\centering
\captionof{table}{
Real-world success rates (\%) for three tasks across three physical embodiments, evaluated over $25$ independent rollouts for each task on each embodiment. \textbf{Bold} indicates the best result.
}
\label{tab:real_world}
\begingroup
\small
\setlength{\tabcolsep}{4pt}
\begin{tabular*}{\textwidth}{@{\extracolsep{\fill}}lccc@{}}
\toprule
 & \multicolumn{3}{c}{Success Rate (\%)} \\
\cmidrule(l){2-4}
 & ACT~\citep{zhao2023act} & GR00T-N1.6~\citep{bjorck2025gr00t} & \textbf{\ourmethod (ours)} \\
\midrule
\multicolumn{4}{@{}l}{\textit{FR3 (single-arm)}} \\
\subrow Kiwi $\rightarrow$ basket & 68 & 84 & \textbf{100} \\
\subrow Pour water & 44 & 76 & \textbf{92} \\
\subrow Book $\rightarrow$ shelf & 36 & 64 & \textbf{80} \\
\midrule
\multicolumn{4}{@{}l}{\textit{COBOT Magic (dual-arm)}} \\
\subrow Kiwi $\rightarrow$ basket & 60 & 80 & \textbf{92} \\
\subrow Pour water & 36 & 64 & \textbf{76} \\
\subrow Book $\rightarrow$ shelf & 28 & 60 & \textbf{80} \\
\midrule
\multicolumn{4}{@{}l}{\textit{G1 (humanoid)}} \\
\subrow Kiwi $\rightarrow$ basket & 20 & 52 & \textbf{72} \\
\subrow Pour water & 0 & 32 & \textbf{52} \\
\subrow Book $\rightarrow$ shelf & 0 & 24 & \textbf{36} \\
\midrule
\textbf{Average} & 32.4 & 59.6 & \textbf{75.6} \\
\bottomrule
\end{tabular*}
\endgroup
\end{minipage}
\end{figure*}

\paragraph{Implementation Details.}
We employ Qwen3-VL-2B~\cite{bai2025qwen3} as the VLM and SANA-600M~\cite{xie2024sana}, initialized from its pretrained weights, as the future generation head. The MoE action head is a 16-layer Diffusion Transformer (DiT) trained from scratch with $K{=}3$ experts, and we use $N{=}96$ shared query tokens.
Each embodiment observes a single frame per camera: two views for the single-arm robots (third-person and wrist), one egocentric view for the GR-1 and G1 humanoids, and three views for the dual-arm robots (egocentric and one per wrist). All views are resized to $256{\times}256$ before entering the VLM. No proprioceptive state is used, and actions are min-max normalized per dimension.
The action horizon $H_{e}$, which also sets the future offset $\Delta_{e}$, is $8$ for the single-arm robots, $16$ for the GR-1 and G1 humanoids, and $50$ for the dual-arm robots.
Stage~1 runs for 100{,}000 steps and Stage~2 for 200{,}000 steps, both with an effective batch size of 512 on 16 H100 GPUs using AdamW.
The future loss weight is $\lambda_{w}{=}0.05$, and inference integrates the action flow with $4$ Euler steps.
For real-world deployment, we jointly finetune the co-trained checkpoint on 1{,}800 demonstrations spanning three tasks across three embodiments for 5{,}000 steps.

\paragraph{Baselines.}
We organize VLA and WAM methods by checkpoint scope, following the grouping in Tables~\ref{tab:main_robotwin},~\ref{tab:main_robocasa}, and~\ref{tab:main_libero}.
(1) \textit{Per-benchmark specialists} are trained, finetuned, or post-trained separately for each benchmark. This group also covers methods pretrained on multi-robot data, because their reported results still come from per-benchmark checkpoints.
(2) \textit{Single-checkpoint generalists} evaluate a single checkpoint on all benchmarks without per-benchmark finetuning.

\subsection{Comparisons with State-of-the-Art Methods (Q1)}
\label{sec:main_results}
Tables~\ref{tab:main_robotwin},~\ref{tab:main_robocasa}, and~\ref{tab:main_libero} report the simulation results.
We evaluate our \ourmethod with a single checkpoint trained on three simulation benchmarks, testing whether one model can handle different robot bodies and action spaces.

With a single checkpoint, our \ourmethod matches or surpasses the strongest methods on all three benchmarks.
On RoboTwin~2.0 with its 14-DoF dual-arm robot, it attains 88.78\% in the clean setting and 89.26\% in the randomized setting, outperforming the concurrent generalist Qwen-VLA by 2.68 and 2.06 points, respectively. Averaged over both settings, it achieves 89.02\%, 2.37 points above Qwen-VLA.
On RoboCasa-GR1 with its 29-DoF humanoid, it achieves 59.25\%, exceeding the strongest per-benchmark specialist ABot-M0 (58.3\%) by 0.95 points, the WAM baseline LDA-1B (55.4\%) by 3.85 points, and the concurrent generalist Qwen-VLA (56.7\%) by 2.55 points.
On LIBERO with its 7-DoF single-arm robot, it reaches 98.0\%, outperforming the WAM baseline Fast-WAM (97.6\%) by 0.4 points and OpenVLA-OFT (97.1\%) by 0.9 points, while trailing the best finetuned X-VLA by only 0.1 points.

\subsection{Real-World Deployment (Q1)}
\label{sec:real_world}

Beyond simulation, we evaluate our \ourmethod on three tabletop manipulation tasks across three physical robots (Fig.~\ref{fig:real_world}). 
The tasks are placing a kiwi into a basket, pouring water from a cup into a bowl, and placing a book onto a shelf.
We jointly finetune the same cross-embodiment checkpoint trained on simulation benchmarks with $200$ teleoperated demonstrations for each task on each embodiment.
LIBERO uses a simulated Franka Emika Panda, whereas our real-world platform is the FR3. 
The two share a closely matched 7-DoF kinematic structure.
The COBOT Magic and G1 similarly reuse the experts trained for the 14-DoF RoboTwin~2.0 and 29-DoF RoboCasa-GR1 embodiments, respectively.

We compare against ACT~\cite{zhao2023act} and the pretrained GR00T-N1.6 generalist~\cite{bjorck2025gr00t} on the same tasks. ACT is trained independently for each task on each embodiment, yielding nine separate checkpoints. 
For GR00T-N1.6, we initialize from the pretrained GR00T-N1.6-3B checkpoint and jointly finetune a single checkpoint on the demonstrations from all three tasks across three embodiments.
We evaluate every policy over $25$ independent rollouts for each task on each embodiment, and Table~\ref{tab:real_world} reports the success rates.
Our real-world evaluation protocol follows prior real-robot studies~\cite{zhao2023act,BlackK-RSS-25}.
ACT averages $32.4\%$ across the nine separate policies, while the jointly finetuned GR00T-N1.6 checkpoint reaches $59.6\%$. Our \ourmethod attains $75.6\%$ with a single jointly finetuned checkpoint, outperforming GR00T-N1.6 by $16.0$ points and demonstrating stronger cross-embodiment real-world control.

\begin{table}[t]
\centering
\caption{
Ablation study on the three core design choices of our \ourmethod.
We report success rates (\%) of a single co-trained checkpoint on all three benchmarks.
\textbf{Bold} indicates the best result, and \underline{underline} indicates the second best.
}
\label{tab:ablation}
\begingroup
\small
\setlength{\tabcolsep}{3pt}
\begin{tabular*}{\textwidth}{@{\extracolsep{\fill}}lccc@{}}
\toprule
& \multicolumn{3}{c}{Success Rate (\%)} \\
\cmidrule(l){2-4}
 & RoboTwin & RoboCasa-GR1 & LIBERO \\
\midrule
\ourmethod (ours) & \textbf{89.02} & \textbf{59.25} & \textbf{98.0} \\
\midrule
\multicolumn{4}{l}{\textit{Future-prediction objective (Q2)}} \\
\subrow w/o future prediction & 86.67 & 56.75 & 96.1 \\
\subrow w/o Stage-1 pretraining & 87.76 & 58.33 & 96.7 \\
\midrule
\multicolumn{4}{l}{\textit{Embodiment-specific MoE action head (Q3)}} \\
\subrow Shared dense head & 87.85 & 57.17 & 96.8 \\
\midrule
\multicolumn{4}{l}{\textit{Embodiment metadata (Q4)}} \\
\subrow w/o metadata & \underline{88.50} & \underline{58.75} & \underline{97.6} \\
\bottomrule
\end{tabular*}
\endgroup
\end{table}

\subsection{Ablation Study (Q2, Q3, Q4)}
\label{sec:ablation}

To evaluate the individual contributions of the future-prediction objective, the embodiment-specific MoE action head, and the embodiment metadata, we train ablated variants on the same data setting with identical hyperparameters and evaluate each with a single checkpoint on all three benchmarks. The results are summarized in Table~\ref{tab:ablation}.

\paragraph{Future Prediction Objective (Q2).}
Removing the future-prediction objective causes the largest drop, 2.4 points on RoboTwin~2.0 and 2.5 points on RoboCasa-GR1, supporting the utility of future supervision for joint policy learning. Removing only Stage-1 pretraining also degrades performance on all three benchmarks.

\paragraph{Embodiment-Specific MoE Action Head (Q3).}
Replacing the MoE action head with a shared dense head costs 1.2 points on RoboTwin~2.0 and 2.1 points on RoboCasa-GR1. This supports the benefit of embodiment-specific action realization for joint cross-embodiment training.

\paragraph{Embodiment Metadata (Q4).}
Removing the embodiment metadata costs up to 0.5 points, a small but consistent drop. The verbalized context does help disambiguate embodiment and data source.

\subsection{Future-Contact Probe (Q2)}
\label{sec:future_contact_probe}

To complement the behavioral ablation for Q2, we probe representation content directly.
A single-layer linear probe tests what information is linearly decodable from the frozen query states $\mathbf{Z}$.
Decoding the future contact sequence isolates whether future supervision makes contact onset and release decodable, rather than whether current contact is already visible in generic vision-language features.

Let $c_t\in\{0,1\}$ be the per-frame LIBERO contact annotation over the action-chunk horizon $H{=}8$.
We define two event-level tasks conditioned on the current state $c_t$. Out of contact ($c_t{=}0$), \texttt{onset} asks whether contact begins within the horizon. In contact ($c_t{=}1$), \texttt{release} asks whether it ends.
Because each subset fixes the current contact state, a trivial predictor that echoes it cannot discriminate transitions, so probe performance must come from anticipating the state change itself.
We report the area under the ROC curve (AUROC) and the area under the precision-recall curve (AUPRC) for each event, together with the probe's class-weighted binary cross-entropy (BCE). The supplementary material gives the full probe protocol.
As shown in Table~\ref{tab:future_contact_probe}, the full model decodes both onset and release markedly better than the variant without future prediction, showing that future supervision enriches the shared query representation with dynamics priors.

\begin{table}[t]
\centering
\caption{
Linear-probe decoding of future contact on LIBERO. We report threshold-free AUROC and AUPRC (\%) for contact onset and release. We also report the probe's class-weighted BCE, which is a loss value rather than a percentage. Future-prediction supervision improves every metric.
}
\label{tab:future_contact_probe}
\begingroup
\small
\begin{tabular*}{\textwidth}{@{\extracolsep{\fill}}lcc@{}}
\toprule
 & w/o future pred. & Full \ourmethod \\
\midrule
\multicolumn{3}{@{}l}{\textit{Metrics (\%)}} \\
\subrow Onset AUROC $\uparrow$ & 95.2 & \textbf{97.3} \\
\subrow Onset AUPRC $\uparrow$ & 70.8 & \textbf{86.3} \\
\subrow Release AUROC $\uparrow$ & 92.3 & \textbf{94.4} \\
\subrow Release AUPRC $\uparrow$ & 64.8 & \textbf{72.8} \\
\midrule
\multicolumn{3}{@{}l}{\textit{Loss}} \\
\subrow Weighted BCE $\downarrow$ & 0.227 & \textbf{0.142} \\
\bottomrule
\end{tabular*}
\endgroup
\end{table}

\section{Conclusion}
\label{sec:conclusion}

We proposed a cross-embodiment learning paradigm for generalist manipulation that learns shared dynamics priors and embodiment-specific control, and instantiated it as our \ourmethod.
This paradigm can be extended to arbitrary embodiments and heterogeneous data.
The three embodiment families instantiated here are a single-arm platform, a dual-arm platform, and a humanoid platform.
To learn shared dynamics priors, we use a future-prediction objective across action-free videos and action-labeled robot demonstrations, supervising a shared query interface to retain regularities of object motion, contact, and scene evolution.
To realize embodiment-specific control, we condition an MoE action head on the shared dynamics priors to generate actions for each embodiment.
With a single co-trained checkpoint, our \ourmethod achieves a 98.0\% success rate on LIBERO, 59.25\% on RoboCasa-GR1, and 89.02\% on RoboTwin~2.0.
Finetuning this checkpoint on real-world demonstrations from three physical embodiments yields a single unified policy that averages 75.6\% success on three tasks across three embodiments.

\clearpage
\appendix
\section*{Supplementary Material}
\setcounter{table}{0}
\setcounter{figure}{0}
\setcounter{equation}{0}
\renewcommand{\thetable}{S\arabic{table}}
\renewcommand{\thefigure}{S\arabic{figure}}
\renewcommand{\theequation}{S\arabic{equation}}
\section{Per-Task Results on RoboCasa-GR1}
\label{sec:appendix_robocasa}

Table~\ref{tab:robocasa_tasks} reports the per-task success rates (SR) of our single co-trained checkpoint on all 24 RoboCasa-GR1~\cite{robocasa,bjorck2025gr00t} tabletop tasks, evaluated with 50 rollouts per task (1{,}200 rollouts in total).
The results are consistent across both task families: 55.3\% on the six pick-and-place-into-articulated-receptacle tasks and 60.6\% on the eighteen container-to-container tasks.

\medskip
\begin{minipage}{\textwidth}
\centering
\captionsetup{hypcap=false}
\captionof{table}{
Per-task success rates (\%) of our \ourmethod on the 24 RoboCasa-GR1 tabletop tasks~\cite{bjorck2025gr00t}, evaluated with 50 rollouts per task.
}
\label{tab:robocasa_tasks}
\begingroup
\small
\begin{tabular*}{\textwidth}{@{\extracolsep{\fill}}lclc@{}}
\toprule
Task & SR (\%) & Task & SR (\%) \\
\midrule
BottleToCabinetClose & 64 & PlacematToBowl & 62 \\
CanToDrawerClose & 70 & PlacematToPlate & 68 \\
CupToDrawerClose & 42 & PlacematToTieredshelf & 34 \\
MilkToMicrowaveClose & 64 & PlateToBowl & 52 \\
PotatoToMicrowaveClose & 32 & PlateToCardboardbox & 62 \\
WineToCabinetClose & 60 & PlateToPan & 70 \\
CuttingboardToBasket & 48 & PlateToPlate & 72 \\
CuttingboardToCardboardbox & 62 & TrayToCardboardbox & 54 \\
CuttingboardToPan & 76 & TrayToPlate & 82 \\
CuttingboardToPot & 74 & TrayToPot & 66 \\
CuttingboardToTieredbasket & 56 & TrayToTieredbasket & 70 \\
PlacematToBasket & 42 & TrayToTieredshelf & 40 \\
\midrule
\multicolumn{3}{l}{\textbf{Average}} & \textbf{59.25} \\
\bottomrule
\end{tabular*}
\endgroup
\end{minipage}

\section{Per-Task Results on RoboTwin~2.0}
\label{sec:appendix_robotwin}

Table~\ref{tab:robotwin_tasks} reports the SR for each task on the RoboTwin~2.0~\cite{robotwin,robotwin2} benchmark with the ALOHA-AgileX embodiment under the clean evaluation setting, using 100 rollouts per task.
The average is 88.78\% over all 50 tasks.
Table~\ref{tab:robotwin_tasks_randomized} reports the corresponding results under the randomized setting, which perturbs backgrounds, lighting, and object placements; the average is 89.26\% over all 50 tasks.
The overall average across both settings is 89.02\%, which is the figure reported in the main paper.

\section{Additional Real-World Qualitative Results}
\label{sec:supp_real_world_qualitative}

\begin{figure*}[t]
\centering
\includegraphics[width=\textwidth]{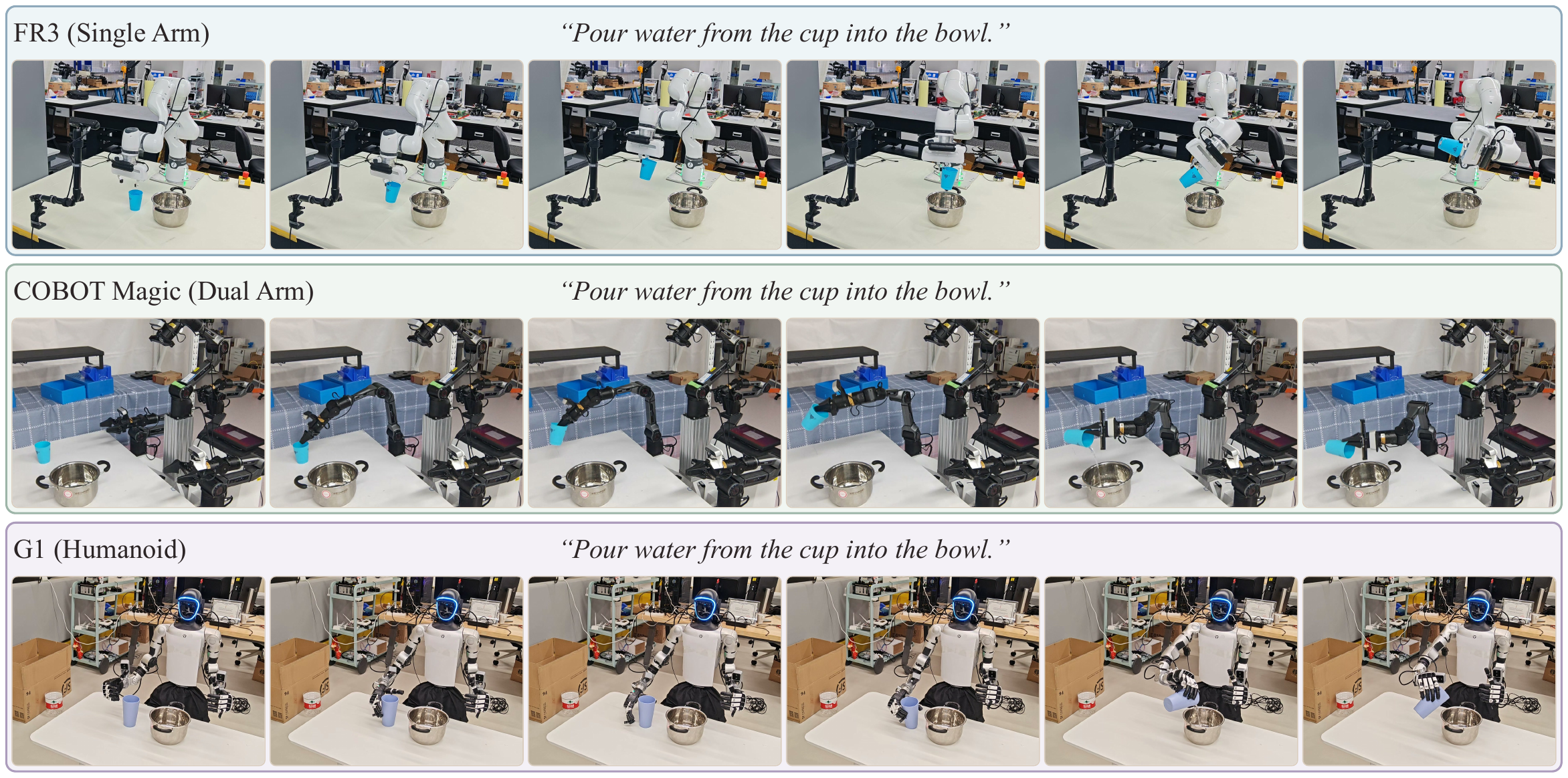}
\caption{
Additional real-world qualitative results for water pouring on FR3, COBOT
Magic, and G1.
Each row shows six chronological frames under the instruction
``Pour water from the cup into the bowl.''
}
\label{fig:supp_real_world_pour_water}
\end{figure*}

\begin{figure*}[t]
\centering
\includegraphics[width=\textwidth]{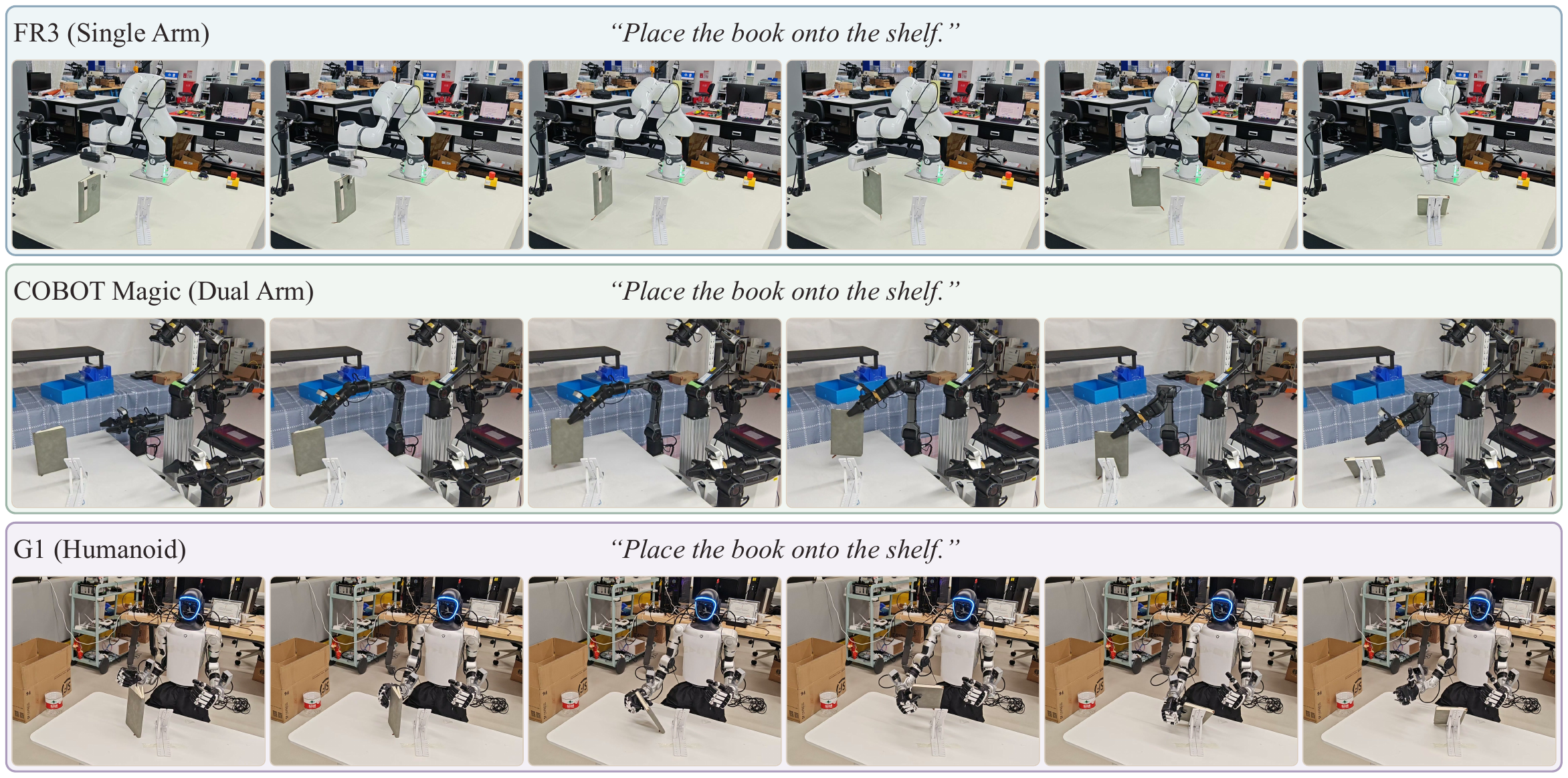}
\caption{
Additional real-world qualitative results for book placement on FR3, COBOT
Magic, and G1.
Each row shows six chronological frames under the instruction
``Place the book onto the shelf.''
}
\label{fig:supp_real_world_book2shelf}
\end{figure*}

Figures~\ref{fig:supp_real_world_pour_water}
and~\ref{fig:supp_real_world_book2shelf} provide additional qualitative
rollouts for the water-pouring and book-placement tasks on all three physical
platforms: FR3, COBOT Magic, and G1.

\begin{table*}[t]
\centering
\caption{
Per-task success rates (\%) on RoboTwin~2.0 (ALOHA-AgileX, clean setting, 100 rollouts per task).
}
\label{tab:robotwin_tasks}
\begingroup
\small
\setlength{\tabcolsep}{5pt}
\begin{tabular*}{\textwidth}{@{\extracolsep{\fill}}lclclc@{}}
\toprule
Task & SR (\%) & Task & SR (\%) & Task & SR (\%) \\
\midrule
Adjust Bottle & 100 & Open Microwave & 100 & Place Object Stand & 98 \\
Beat Block Hammer & 95 & Pick Diverse Bottles & 96 & Place Phone Stand & 100 \\
Blocks Ranking RGB & 96 & Pick Dual Bottles & 96 & Place Shoe & 100 \\
Blocks Ranking Size & 63 & Place A2B Left & 98 & Press Stapler & 97 \\
Click Alarmclock & 99 & Place A2B Right & 98 & Put Bottles Dustbin & 72 \\
Click Bell & 97 & Place Bread Basket & 71 & Put Object Cabinet & 90 \\
Dump Bin Bigbin & 94 & Place Bread Skillet & 96 & Rotate QR Code & 96 \\
Grab Roller & 98 & Place Burger Fries & 99 & Scan Object & 96 \\
Handover Block & 75 & Place Can Basket & 58 & Shake Bottle & 100 \\
Handover Mic & 98 & Place Cans Plasticbox & 96 & Shake Bottle Horizontally & 97 \\
Hanging Mug & 35 & Place Container Plate & 94 & Stack Blocks Three & 51 \\
Lift Pot & 93 & Place Dual Shoes & 91 & Stack Blocks Two & 95 \\
Move Can Pot & 77 & Place Empty Cup & 95 & Stack Bowls Three & 70 \\
Move Pillbottle Pad & 100 & Place Fan & 95 & Stack Bowls Two & 90 \\
Move Playingcard Away & 97 & Place Mouse Pad & 76 & Stamp Seal & 83 \\
Move Stapler Pad & 70 & Place Object Basket & 91 & Turn Switch & 88 \\
Open Laptop & 99 & Place Object Scale & 80 &  &  \\
\midrule
\multicolumn{4}{l}{\textbf{Average}} & & \textbf{88.78} \\
\bottomrule
\end{tabular*}
\endgroup
\end{table*}

\begin{table*}[t]
\centering
\caption{
Per-task success rates (\%) on RoboTwin~2.0 (ALOHA-AgileX, randomized setting, 100 rollouts per task).
}
\label{tab:robotwin_tasks_randomized}
\begingroup
\small
\setlength{\tabcolsep}{5pt}
\begin{tabular*}{\textwidth}{@{\extracolsep{\fill}}lclclc@{}}
\toprule
Task & SR (\%) & Task & SR (\%) & Task & SR (\%) \\
\midrule
Adjust Bottle & 100 & Open Microwave & 100 & Place Object Stand & 96 \\
Beat Block Hammer & 93 & Pick Diverse Bottles & 86 & Place Phone Stand & 98 \\
Blocks Ranking RGB & 94 & Pick Dual Bottles & 86 & Place Shoe & 97 \\
Blocks Ranking Size & 58 & Place A2B Left & 96 & Press Stapler & 98 \\
Click Alarmclock & 95 & Place A2B Right & 96 & Put Bottles Dustbin & 65 \\
Click Bell & 95 & Place Bread Basket & 83 & Put Object Cabinet & 84 \\
Dump Bin Bigbin & 98 & Place Bread Skillet & 97 & Rotate QR Code & 85 \\
Grab Roller & 98 & Place Burger Fries & 94 & Scan Object & 80 \\
Handover Block & 69 & Place Can Basket & 74 & Shake Bottle & 100 \\
Handover Mic & 98 & Place Cans Plasticbox & 98 & Shake Bottle Horizontally & 99 \\
Hanging Mug & 36 & Place Container Plate & 94 & Stack Blocks Three & 77 \\
Lift Pot & 98 & Place Dual Shoes & 92 & Stack Blocks Two & 94 \\
Move Can Pot & 82 & Place Empty Cup & 98 & Stack Bowls Three & 83 \\
Move Pillbottle Pad & 100 & Place Fan & 97 & Stack Bowls Two & 96 \\
Move Playingcard Away & 96 & Place Mouse Pad & 71 & Stamp Seal & 93 \\
Move Stapler Pad & 63 & Place Object Basket & 100 & Turn Switch & 99 \\
Open Laptop & 100 & Place Object Scale & 84 &  &  \\
\midrule
\multicolumn{4}{l}{\textbf{Average}} & & \textbf{89.26} \\
\bottomrule
\end{tabular*}
\endgroup
\end{table*}

\clearpage
\begin{thebibliography}{63}
\providecommand{\natexlab}[1]{#1}
\providecommand{\url}[1]{\texttt{#1}}
\expandafter\ifx\csname urlstyle\endcsname\relax
  \providecommand{\doi}[1]{doi: #1}\else
  \providecommand{\doi}{doi: \begingroup \urlstyle{rm}\Url}\fi

\bibitem[Bai et~al.(2025)Bai, Cai, Chen, Chen, Chen, Cheng, Deng, Ding, Gao,
  et~al.]{bai2025qwen3}
Shuai Bai, Yuxuan Cai, Ruizhe Chen, Keqin Chen, Xionghui Chen, Zesen Cheng,
  Lianghao Deng, Wei Ding, Chang Gao, et~al.
\newblock {Qwen3-VL} technical report.
\newblock \emph{arXiv preprint arXiv:2511.21631}, 2025.

\bibitem[Bi et~al.(2025)Bi, Tan, Xie, et~al.]{motus}
Hongzhe Bi, Hengkai Tan, Shenghao Xie, et~al.
\newblock Motus: A unified latent action world model.
\newblock \emph{arXiv preprint arXiv:2512.13030}, 2025.

\bibitem[Bjorck et~al.(2025)Bjorck, Casta{\~n}eda, Cherniadev, Da, Ding, Fan,
  Fang, Fox, Hu, Huang, et~al.]{bjorck2025gr00t}
Johan Bjorck, Fernando Casta{\~n}eda, Nikita Cherniadev, Xingye Da, Runyu Ding,
  Linxi Fan, Yu~Fang, Dieter Fox, Fengyuan Hu, Spencer Huang, et~al.
\newblock {GR00T N1}: An open foundation model for generalist humanoid robots.
\newblock \emph{arXiv preprint arXiv:2503.14734}, 2025.

\bibitem[Black et~al.(2025{\natexlab{a}})Black, Brown, Darpinian, Dhabalia,
  Driess, Esmail, Equi, et~al.]{black2025pi}
Kevin Black, Noah Brown, James Darpinian, Karan Dhabalia, Danny Driess, Adnan
  Esmail, Michael~Robert Equi, et~al.
\newblock $\pi_{0.5}$: a vision-language-action model with open-world
  generalization.
\newblock In \emph{9th Annual Conference on Robot Learning},
  2025{\natexlab{a}}.

\bibitem[Black et~al.(2025{\natexlab{b}})Black, Brown, Driess, Esmail, Equi,
  Finn, et~al.]{BlackK-RSS-25}
Kevin Black, Noah Brown, Danny Driess, Adnan Esmail, Michael~Robert Equi,
  Chelsea Finn, et~al.
\newblock $\pi_{0}$: A vision-language-action flow model for general robot
  control.
\newblock In \emph{Proceedings of Robotics: Science and Systems}, Los Angeles,
  CA, USA, June 2025{\natexlab{b}}.
\newblock \doi{10.15607/RSS.2025.XXI.010}.

\bibitem[Brohan et~al.(2023)Brohan, Brown, Carbajal, Chebotar, Dabis, Finn,
  Gopalakrishnan, Hausman, Herzog, Hsu, et~al.]{brohan2023rt}
Anthony Brohan, Noah Brown, Justice Carbajal, Yevgen Chebotar, Joseph Dabis,
  Chelsea Finn, Keerthana Gopalakrishnan, Karol Hausman, Alexander Herzog,
  Jasmine Hsu, et~al.
\newblock Rt-1: Robotics transformer for real-world control at scale.
\newblock \emph{Robotics: Science and Systems XIX}, 2023.

\bibitem[Bu et~al.(2025)Bu, Yang, Cai, Gao, Ren, Yao, Luo, and Li]{univla}
Qingwen Bu, Yanting Yang, Jisong Cai, Shenyuan Gao, Guanghui Ren, Maoqing Yao,
  Ping Luo, and Hongyang Li.
\newblock Univla: Learning to act anywhere with task-centric latent actions.
\newblock \emph{arXiv preprint arXiv:2505.06111}, 2025.

\bibitem[Cen et~al.(2025)Cen, Yu, Yuan, et~al.]{worldvla}
Jun Cen, Chaohui Yu, Hangjie Yuan, et~al.
\newblock Worldvla: Towards autoregressive action world model.
\newblock \emph{arXiv preprint arXiv:2506.21539}, 2025.

\bibitem[Chen et~al.(2026)Chen, Song, Ding, Zhou, Zhao, Tang, Wang, and
  Li]{chen2026unified}
Jiayi Chen, Wenxuan Song, Pengxiang Ding, Ziyang Zhou, Han Zhao, Barrett Tang,
  Donglin Wang, and Haoang Li.
\newblock Unified diffusion vla: Vision-language-action model via joint
  discrete denosing diffusion process.
\newblock In \emph{International Conference on Learning Representations},
  volume 2026, pages 139291--139311, 2026.

\bibitem[Chen et~al.(2025{\natexlab{a}})Chen, Chen, Chen, Cai, Liu, Liang, Li,
  Lin, Ge, Gu, et~al.]{robotwin2}
Tianxing Chen, Zanxin Chen, Baijun Chen, Zijian Cai, Yibin Liu, Qiwei Liang,
  Zixuan Li, Xianliang Lin, Yiheng Ge, Zhenyu Gu, et~al.
\newblock Robotwin 2.0: A scalable data generator and benchmark with strong
  domain randomization for robust bimanual robotic manipulation.
\newblock \emph{arXiv preprint arXiv:2506.18088}, 2025{\natexlab{a}}.

\bibitem[Chen et~al.(2025{\natexlab{b}})Chen, Ge, Tang, et~al.]{moto}
Yi~Chen, Yuying Ge, Weiliang Tang, et~al.
\newblock Moto: Latent motion token as the bridging language for learning robot
  manipulation from videos.
\newblock In \emph{ICCV}, 2025{\natexlab{b}}.

\bibitem[Chi et~al.(2025)Chi, Xu, Feng, Cousineau, Du, Burchfiel, Tedrake, and
  Song]{chi2025diffusion}
Cheng Chi, Zhenjia Xu, Siyuan Feng, Eric Cousineau, Yilun Du, Benjamin
  Burchfiel, Russ Tedrake, and Shuran Song.
\newblock Diffusion policy: Visuomotor policy learning via action diffusion.
\newblock \emph{The International Journal of Robotics Research}, 44\penalty0
  (10-11):\penalty0 1684--1704, 2025.

\bibitem[Doshi et~al.(2024)Doshi, Walke, Mees, et~al.]{crossformer}
Ria Doshi, Homer Walke, Oier Mees, et~al.
\newblock Scaling cross-embodied learning: One policy for manipulation,
  navigation, locomotion and aviation.
\newblock In \emph{Conference on Robot Learning}, 2024.

\bibitem[Du et~al.(2025)Du, Liu, Liang, Shen, Cao, Zheng, Feng, Wu, Yang, and
  Jiang]{himoevla}
Zhiying Du, Bei Liu, Yaobo Liang, Yichao Shen, Haidong Cao, Xiangyu Zheng,
  Zhiyuan Feng, Zuxuan Wu, Jiaolong Yang, and Yu-Gang Jiang.
\newblock Himoe-vla: Hierarchical mixture-of-experts for generalist
  vision-language-action policies.
\newblock \emph{arXiv preprint arXiv:2512.05693}, 2025.

\bibitem[{Gemini Robotics Team}(2025)]{gemini15}
{Gemini Robotics Team}.
\newblock Gemini robotics 1.5: Pushing the frontier of generalist robots with
  advanced embodied reasoning, thinking, and motion transfer.
\newblock \emph{arXiv preprint arXiv:2510.03342}, 2025.

\bibitem[Guo et~al.(2024)Guo, Hu, Zhang, et~al.]{pad}
Yanjiang Guo, Yucheng Hu, Jianke Zhang, et~al.
\newblock Prediction with action: Visual policy learning via joint denoising
  process.
\newblock In \emph{NeurIPS}, 2024.

\bibitem[Hoque et~al.(2025)Hoque, Huang, Yoon, Sivapurapu, and
  Zhang]{hoque2025egodex}
Ryan Hoque, Peide Huang, David~J Yoon, Mouli Sivapurapu, and Jian Zhang.
\newblock Egodex: Learning dexterous manipulation from large-scale egocentric
  video.
\newblock \emph{arXiv preprint arXiv:2505.11709}, 2025.

\bibitem[Hu et~al.(2025)Hu, Guo, Wang, et~al.]{vpp}
Yucheng Hu, Yanjiang Guo, Pengchao Wang, et~al.
\newblock Video prediction policy: A generalist robot policy with predictive
  visual representations.
\newblock In \emph{ICML}, 2025.

\bibitem[Kim et~al.(2024)Kim, Pertsch, Karamcheti, Xiao, Balakrishna, Nair,
  Rafailov, Foster, Sanketi, Vuong, et~al.]{kim2024openvla}
Moo~Jin Kim, Karl Pertsch, Siddharth Karamcheti, Ted Xiao, Ashwin Balakrishna,
  Suraj Nair, Rafael Rafailov, Ethan~P Foster, Pannag~R Sanketi, Quan Vuong,
  et~al.
\newblock Open{VLA}: An open-source vision-language-action model.
\newblock In \emph{8th Annual Conference on Robot Learning}, 2024.

\bibitem[Kim et~al.(2025)Kim, Finn, Liang, et~al.]{KimM1-RSS-25}
Moo~Jin Kim, Chelsea Finn, Percy Liang, et~al.
\newblock {Fine-Tuning Vision-Language-Action Models: Optimizing Speed and
  Success}.
\newblock In \emph{Proceedings of Robotics: Science and Systems}, Los Angeles,
  CA, USA, June 2025.
\newblock \doi{10.15607/RSS.2025.XXI.017}.

\bibitem[Li et~al.(2026{\natexlab{a}})Li, Xu, Yuan, Xu, Karlsson, Zhao, Li, and
  Lu]{xdiffvla}
Boyu Li, Chaoyi Xu, Haoqi Yuan, Xinrun Xu, B{\"o}rje~F. Karlsson, Dongbin Zhao,
  Haoran Li, and Zongqing Lu.
\newblock X-diffvla: X-embodied diffusion action heads for
  vision-language-action models.
\newblock \emph{arXiv preprint arXiv:2605.25044}, 2026{\natexlab{a}}.

\bibitem[Li et~al.(2026{\natexlab{b}})Li, Song, Zhao, Wang, Ding, Wang, Zeng,
  and Li]{li2026spatial}
Fuhao Li, Wenxuan Song, Han Zhao, Jingbo Wang, Pengxiang Ding, Donglin Wang,
  Long Zeng, and Haoang Li.
\newblock Spatial forcing: Implicit spatial representation alignment for
  vision-language-action model.
\newblock In \emph{International Conference on Learning Representations},
  volume 2026, pages 132324--132345, 2026{\natexlab{b}}.

\bibitem[Li et~al.(2025)Li, Gao, Sadigh, and Song]{uva}
Shuang Li, Yihuai Gao, Dorsa Sadigh, and Shuran Song.
\newblock Unified video action model.
\newblock In \emph{Proceedings of Robotics: Science and Systems}, 2025.

\bibitem[Liao et~al.(2025)Liao, Zhou, Huang, et~al.]{genieenvisioner}
Yue Liao, Pengfei Zhou, Siyuan Huang, et~al.
\newblock Genie envisioner: A unified world foundation platform for robotic
  manipulation.
\newblock \emph{arXiv preprint arXiv:2508.05635}, 2025.

\bibitem[Lipman et~al.(2022)Lipman, Chen, Ben-Hamu, Nickel, and
  Le]{lipman2022flow}
Yaron Lipman, Ricky~TQ Chen, Heli Ben-Hamu, Maximilian Nickel, and Matt Le.
\newblock Flow matching for generative modeling.
\newblock \emph{arXiv preprint arXiv:2210.02747}, 2022.

\bibitem[Liu et~al.(2023)Liu, Zhu, Gao, Feng, Liu, Zhu, and
  Stone]{liu2023libero}
Bo~Liu, Yifeng Zhu, Chongkai Gao, Yihao Feng, Qiang Liu, Yuke Zhu, and Peter
  Stone.
\newblock {LIBERO}: Benchmarking knowledge transfer for lifelong robot
  learning.
\newblock \emph{Advances in Neural Information Processing Systems},
  36:\penalty0 44776--44791, 2023.

\bibitem[Liu et~al.(2024)Liu, Wu, Li, Tan, Chen, Wang, Xu, Su, and Zhu]{rdt1b}
Songming Liu, Lingxuan Wu, Bangguo Li, Hengkai Tan, Huayu Chen, Zhengyi Wang,
  Ke~Xu, Hang Su, and Jun Zhu.
\newblock Rdt-1b: A diffusion foundation model for bimanual manipulation.
\newblock \emph{arXiv preprint arXiv:2410.07864}, 2024.

\bibitem[Liu et~al.(2026)Liu, Li, Ma, et~al.]{rdt2}
Songming Liu, Bangguo Li, Kai Ma, et~al.
\newblock Rdt2: Exploring the scaling limit of umi data towards zero-shot
  cross-embodiment generalization.
\newblock \emph{arXiv preprint arXiv:2602.03310}, 2026.

\bibitem[Luo et~al.(2025)Luo, Feng, Zhang, et~al.]{beingh0}
Hao Luo, Yicheng Feng, Wanpeng Zhang, et~al.
\newblock Being-h0: Vision-language-action pretraining from large-scale human
  videos.
\newblock \emph{arXiv preprint arXiv:2507.15597}, 2025.

\bibitem[Lv et~al.(2025)Lv, Kong, Li, Zeng, Qiu, Qu, Song, Chen, Deng, and
  Pang]{lv2025f1}
Qi~Lv, Weijie Kong, Hao Li, Jia Zeng, Zherui Qiu, Delin Qu, Haoming Song, Qizhi
  Chen, Xiang Deng, and Jiangmiao Pang.
\newblock F1: A vision-language-action model bridging understanding and
  generation to actions.
\newblock \emph{arXiv preprint arXiv:2509.06951}, 2025.

\bibitem[Lyu et~al.(2026)Lyu, Liu, Zhang, Liao, Feng, Zhu, Shen, Chen, Zhang,
  Dong, et~al.]{lda1b}
Jiangran Lyu, Kai Liu, Xuheng Zhang, Haoran Liao, Yusen Feng, Wenxuan Zhu,
  Tingrui Shen, Jiayi Chen, Jiazhao Zhang, Yifei Dong, et~al.
\newblock Lda-1b: Scaling latent dynamics action model via universal embodied
  data ingestion.
\newblock \emph{arXiv preprint arXiv:2602.12215}, 2026.

\bibitem[Ma et~al.(2026)Ma, Zheng, Wang, Jiang, Cui, Liang, and Yang]{dit4dit}
Teli Ma, Jia Zheng, Zifan Wang, Chunli Jiang, Andy Cui, Junwei Liang, and Shuo
  Yang.
\newblock Dit4dit: Jointly modeling video dynamics and actions for
  generalizable robot control.
\newblock \emph{arXiv preprint arXiv:2603.10448}, 2026.

\bibitem[Mu et~al.(2025)Mu, Chen, Chen, et~al.]{robotwin}
Yao Mu, Tianxing Chen, Zanxin Chen, et~al.
\newblock Robotwin: Dual-arm robot benchmark with generative digital twins.
\newblock In \emph{Proceedings of the IEEE/CVF Conference on Computer Vision
  and Pattern Recognition}, 2025.

\bibitem[Nasiriany et~al.(2024)Nasiriany, Maddukuri, Zhang, Parikh, Lo, Joshi,
  Mandlekar, and Zhu]{robocasa}
Soroush Nasiriany, Abhiram Maddukuri, Lance Zhang, Adeet Parikh, Aaron Lo,
  Abhishek Joshi, Ajay Mandlekar, and Yuke Zhu.
\newblock Robocasa: Large-scale simulation of everyday tasks for generalist
  robots.
\newblock In \emph{Proceedings of Robotics: Science and Systems}, 2024.

\bibitem[{Octo Model Team} et~al.(2024){Octo Model Team}, Ghosh, Walke,
  et~al.]{octo}
{Octo Model Team}, Dibya Ghosh, Homer Walke, et~al.
\newblock Octo: An open-source generalist robot policy.
\newblock \emph{arXiv preprint arXiv:2405.12213}, 2024.

\bibitem[{Open X-Embodiment Collaboration} et~al.(2023){Open X-Embodiment
  Collaboration}, O'Neill, Rehman, Gupta, Maddukuri, Gupta, Padalkar, Lee,
  Pooley, Gupta, et~al.]{openxembodiment}
{Open X-Embodiment Collaboration}, Abby O'Neill, Abdul Rehman, Abhinav Gupta,
  Abhiram Maddukuri, Abhishek Gupta, Abhishek Padalkar, Abraham Lee, Acorn
  Pooley, Agrim Gupta, et~al.
\newblock Open x-embodiment: Robotic learning datasets and rt-x models.
\newblock \emph{arXiv preprint arXiv:2310.08864}, 2023.

\bibitem[Peebles and Xie(2023)]{peebles2023scalable}
William Peebles and Saining Xie.
\newblock Scalable diffusion models with transformers.
\newblock In \emph{Proceedings of the IEEE/CVF international conference on
  computer vision}, pages 4195--4205, 2023.

\bibitem[Pertsch et~al.(2025)Pertsch, Stachowicz, Ichter, Driess, Nair, Vuong,
  Mees, Finn, and Levine]{PertschK-RSS-25}
Karl Pertsch, Kyle Stachowicz, Brian Ichter, Danny Driess, Suraj Nair, Quan
  Vuong, Oier Mees, Chelsea Finn, and Sergey Levine.
\newblock {FAST: Efficient Action Tokenization for Vision-Language-Action
  Models}.
\newblock In \emph{Proceedings of Robotics: Science and Systems}, Los Angeles,
  CA, USA, June 2025.
\newblock \doi{10.15607/RSS.2025.XXI.012}.

\bibitem[{Qwen Team}(2026)]{qwenvla2026}
{Qwen Team}.
\newblock Qwen-vla: Unifying vision-language-action modeling across tasks,
  environments, and robot embodiments.
\newblock \emph{arXiv preprint arXiv:2605.30280}, 2026.

\bibitem[Song et~al.(2025)Song, Chen, Ding, Zhao, Zhao, Zhong, Ge, Li, Wang,
  Wang, et~al.]{song2025pd}
Wenxuan Song, Jiayi Chen, Pengxiang Ding, Han Zhao, Wei Zhao, Zhide Zhong,
  Zongyuan Ge, Zhijun Li, Donglin Wang, Lujia Wang, et~al.
\newblock Pd-vla: Accelerating vision-language-action model integrated with
  action chunking via parallel decoding.
\newblock In \emph{2025 IEEE/RSJ International Conference on Intelligent Robots
  and Systems (IROS)}, pages 13162--13169. IEEE, 2025.

\bibitem[Song et~al.(2026)Song, Zhou, Zhao, Chen, Ding, Yan, Huang, Tang, Wang,
  and Li]{song2026reconvla}
Wenxuan Song, Ziyang Zhou, Han Zhao, Jiayi Chen, Pengxiang Ding, Haodong Yan,
  Yuxin Huang, Feilong Tang, Donglin Wang, and Haoang Li.
\newblock Reconvla: Reconstructive vision-language-action model as effective
  robot perceiver.
\newblock In \emph{Proceedings of the AAAI Conference on Artificial
  Intelligence}, volume~40, pages 18549--18557, 2026.

\bibitem[starVLA Contributors(2025)]{starvla2025}
starVLA Contributors.
\newblock Starvla: A lego-like codebase for vision-language-action model
  developing.
\newblock GitHub repository, 1 2025.

\bibitem[Tian et~al.(2025)Tian, Yang, Zeng, et~al.]{seer}
Yang Tian, Sizhe Yang, Jia Zeng, et~al.
\newblock Predictive inverse dynamics models are scalable learners for robotic
  manipulation.
\newblock In \emph{ICLR}, 2025.

\bibitem[Wang et~al.(2024)Wang, Chen, Zhao, and He]{hpt}
Lirui Wang, Xinlei Chen, Jialiang Zhao, and Kaiming He.
\newblock Scaling proprioceptive-visual learning with heterogeneous pretrained
  transformers.
\newblock In \emph{NeurIPS}, 2024.

\bibitem[Wu et~al.(2024)Wu, Jing, Cheang, et~al.]{gr1}
Hongtao Wu, Ya~Jing, Chilam Cheang, et~al.
\newblock Unleashing large-scale video generative pretraining for visual robot
  manipulation.
\newblock In \emph{ICLR}, 2024.

\bibitem[Xie et~al.(2024)Xie, Chen, Chen, Cai, Tang, Lin, Zhang, Li, Zhu, Lu,
  and Han]{xie2024sana}
Enze Xie, Junsong Chen, Junyu Chen, Han Cai, Haotian Tang, Yujun Lin, Zhekai
  Zhang, Muyang Li, Ligeng Zhu, Yao Lu, and Song Han.
\newblock Sana: Efficient high-resolution image synthesis with linear diffusion
  transformers.
\newblock \emph{arXiv preprint arXiv:2410.10629}, 2024.

\bibitem[Yan et~al.(2026)Yan, Zhong, Zhu, He, Yuan, Song, Gong, Cai, Zhao, Yan,
  Liu, Chen, and Li]{yan2026svam}
Haodong Yan, Zhide Zhong, Jiaguan Zhu, Junjie He, Weilin Yuan, Wenxuan Song,
  Xin Gong, Yingjie Cai, Guanyi Zhao, Xu~Yan, Bingbing Liu, Ying-Cong Chen, and
  Haoang Li.
\newblock {S-VAM}: Shortcut video-action model by self-distilling geometric and
  semantic foresight.
\newblock \emph{arXiv preprint arXiv:2603.16195}, 2026.

\bibitem[Yang et~al.(2025)Yang, Yu, Wu, et~al.]{egovla}
Ruihan Yang, Qinxi Yu, Yecheng Wu, et~al.
\newblock Egovla: Learning vision-language-action models from egocentric human
  videos.
\newblock \emph{arXiv preprint arXiv:2507.12440}, 2025.

\bibitem[Yang et~al.(2026{\natexlab{a}})Yang, Zeng, Lin, et~al.]{abotm0}
Yandan Yang, Shuang Zeng, Tong Lin, et~al.
\newblock Abot-m0: Vla foundation model for robotic manipulation with action
  manifold learning.
\newblock \emph{arXiv preprint arXiv:2602.11236}, 2026{\natexlab{a}}.

\bibitem[Yang et~al.(2026{\natexlab{b}})Yang, Li, Chen, et~al.]{mantis}
Yi~Yang, Xueqi Li, Yiyang Chen, et~al.
\newblock Mantis: A versatile vision-language-action model with disentangled
  visual foresight.
\newblock In \emph{CVPR}, 2026{\natexlab{b}}.

\bibitem[Yang et~al.(2026{\natexlab{c}})Yang, Liu, Kou, et~al.]{wla0}
Yi~Yang, Zhihong Liu, Siqi Kou, et~al.
\newblock World-language-action model for unified world modeling, language
  reasoning, and action synthesis.
\newblock \emph{arXiv preprint arXiv:2606.05979}, 2026{\natexlab{c}}.

\bibitem[Ye et~al.(2025)Ye, Jang, Jeon, et~al.]{lapa}
Seonghyeon Ye, Joel Jang, Byeongguk Jeon, et~al.
\newblock Latent action pretraining from videos.
\newblock In \emph{ICLR}, 2025.

\bibitem[Yuan et~al.(2026)Yuan, Dong, Liu, and Zhao]{yuan2026fastwam}
Tianyuan Yuan, Zibin Dong, Yicheng Liu, and Hang Zhao.
\newblock Fast-wam: Do world action models need test-time future imagination?
\newblock \emph{arXiv preprint arXiv:2603.16666}, 2026.
\newblock \url{https://arxiv.org/abs/2603.16666}.

\bibitem[Zhang et~al.(2025)Zhang, Liu, Qi, Wang, Yu, Zhang, Dong, He, Wang,
  Zhang, et~al.]{zhangdreamvla}
Wenyao Zhang, Hongsi Liu, Zekun Qi, Yunnan Wang, XinQiang Yu, Jiazhao Zhang,
  Runpei Dong, Jiawei He, He~Wang, Zhizheng Zhang, et~al.
\newblock Dreamvla: A vision-language-action model dreamed with comprehensive
  world knowledge.
\newblock In \emph{The Thirty-ninth Annual Conference on Neural Information
  Processing Systems}, 2025.

\bibitem[Zhao et~al.(2025)Zhao, Lu, Kim, Fu, Zhang, Wu, Li, Ma, Han, Finn,
  et~al.]{zhao2025cot}
Qingqing Zhao, Yao Lu, Moo~Jin Kim, Zipeng Fu, Zhuoyang Zhang, Yecheng Wu,
  Zhaoshuo Li, Qianli Ma, Song Han, Chelsea Finn, et~al.
\newblock {CoT-VLA}: Visual chain-of-thought reasoning for
  vision-language-action models.
\newblock In \emph{Proceedings of the Computer Vision and Pattern Recognition
  Conference}, pages 1702--1713, 2025.

\bibitem[Zhao et~al.(2023)Zhao, Kumar, Levine, and Finn]{zhao2023act}
Tony~Z Zhao, Vikash Kumar, Sergey Levine, and Chelsea Finn.
\newblock Learning fine-grained bimanual manipulation with low-cost hardware.
\newblock In \emph{Robotics: Science and Systems (RSS)}, 2023.

\bibitem[Zheng et~al.(2025{\natexlab{a}})Zheng, Li, Liu, et~al.]{uniact}
Jinliang Zheng, Jianxiong Li, Dongxiu Liu, et~al.
\newblock Universal actions for enhanced embodied foundation models.
\newblock In \emph{CVPR}, 2025{\natexlab{a}}.

\bibitem[Zheng et~al.(2025{\natexlab{b}})Zheng, Li, Wang, Liu, Kang, Feng,
  Zheng, Zou, Chen, Zeng, et~al.]{zheng2025x}
Jinliang Zheng, Jianxiong Li, Zhihao Wang, Dongxiu Liu, Xirui Kang, Yuchun
  Feng, Yinan Zheng, Jiayin Zou, Yilun Chen, Jia Zeng, et~al.
\newblock {X-VLA}: Soft-prompted transformer as scalable cross-embodiment
  vision-language-action model.
\newblock \emph{arXiv preprint arXiv:2510.10274}, 2025{\natexlab{b}}.

\bibitem[Zheng et~al.(2025{\natexlab{c}})Zheng, Wang, Reed, et~al.]{flare}
Ruijie Zheng, Jing Wang, Scott Reed, et~al.
\newblock Flare: Robot learning with implicit world modeling.
\newblock \emph{arXiv preprint arXiv:2505.15659}, 2025{\natexlab{c}}.

\bibitem[Zhong et~al.(2025)Zhong, Yan, Li, Liu, Gong, Zhang, Song, Chen, Zheng,
  Wang, et~al.]{zhong2025flowvla}
Zhide Zhong, Haodong Yan, Junfeng Li, Xiangchen Liu, Xin Gong, Tianran Zhang,
  Wenxuan Song, Jiayi Chen, Xinhu Zheng, Hesheng Wang, et~al.
\newblock {FlowVLA}: Visual chain of thought-based motion reasoning for
  vision-language-action models.
\newblock \emph{arXiv preprint arXiv:2508.18269}, 2025.

\bibitem[Zhong et~al.(2026)Zhong, Li, He, et~al.]{zhong2026dualcotvla}
Zhide Zhong, Junfeng Li, Junjie He, et~al.
\newblock Dualcot-vla: Visual-linguistic chain of thought via parallel
  reasoning for vision-language-action models.
\newblock \emph{arXiv preprint arXiv:2603.22280}, 2026.

\bibitem[Zhu et~al.(2025)Zhu, Yu, Feng, Burchfiel, Shah, and Gupta]{uwm}
Chuning Zhu, Raymond Yu, Siyuan Feng, Benjamin Burchfiel, Paarth Shah, and
  Abhishek Gupta.
\newblock Unified world models: Coupling video and action diffusion for
  pretraining on large robotic datasets.
\newblock \emph{arXiv preprint arXiv:2504.02792}, 2025.

\bibitem[Zitkovich et~al.(2023)Zitkovich, Yu, Xu, Xu, Xiao, Xia, Wu, Wohlhart,
  Welker, Wahid, et~al.]{zitkovich2023rt}
Brianna Zitkovich, Tianhe Yu, Sichun Xu, Peng Xu, Ted Xiao, Fei Xia, Jialin Wu,
  Paul Wohlhart, Stefan Welker, Ayzaan Wahid, et~al.
\newblock {RT-2}: Vision-language-action models transfer web knowledge to
  robotic control.
\newblock In \emph{Conference on Robot Learning}, pages 2165--2183. PMLR, 2023.

\end{thebibliography}
\end{document}